\documentclass[runningheads]{llncs}
\usepackage[T1]{fontenc}

\usepackage{mwe}
\usepackage{graphicx}
\usepackage{lipsum}
\usepackage{booktabs}
\usepackage{amsmath,amssymb,amsfonts}
\usepackage{algpseudocode}
\usepackage{algorithm}
\usepackage{algorithmicx}
\algblock{Input}{EndInput}
\algnotext{EndInput}
\algblock{Output}{EndOutput}
\algnotext{EndOutput}
\usepackage{multirow}

\usepackage{array}

\usepackage{graphicx}
\usepackage{textcomp}
\usepackage{xcolor}
\usepackage{amsmath}
\usepackage{graphics} 
\usepackage{epsfig} 
\usepackage{mathptmx} 
\usepackage{times} 
\usepackage{amsmath} 
\usepackage{amssymb}  
\usepackage{newtxtext, newtxmath}
\usepackage{ulem}
\usepackage{csquotes}
\usepackage[english]{babel}
\usepackage[square,sort,comma,numbers]{natbib}
\usepackage{fontawesome5}
\begin{document}
\title{A* search algorithm for an optimal investment problem in vehicle-sharing systems}

%
%
\author{Ba Luat Le \inst{1} \orcidID{0000-0002-1980-4274}
 \and
Layla Martin \inst{2} \orcidID{0000-0002-1264-0457} \and
Emrah Demir \inst{3} \orcidID{0000-0002-4726-2556}
\and
Duc Minh Vu \faIcon{envelope} \inst{1} \orcidID{0000-0001-5882-3868} 
}
\authorrunning{Le et al.}
%
\institute{ORLab \& Faculty of Computer Science, Phenikaa University, Hanoi, Vietnam\\
\faIcon{envelope} Corresponding author: minh.vuduc@phenikaa-uni.edu.vn
 \and
Department of Industrial Engineering \& Eindhoven AI Systems Institute, Eindhoven University of Technology\\
\and 
Cardiff Business School, Cardiff University\\
}
\maketitle              
\begin{abstract}
 We study an optimal investment problem that arises in the context of the vehicle-sharing system. Given a set of locations to build stations, we need to determine $i$) the sequence of stations to be built and the number of vehicles to acquire in order to obtain the target state where all stations are built, and $ii$) the number of vehicles to acquire and their allocation in order to maximize the total profit returned by operating the system when some or all stations are open. The profitability associated with operating open stations, measured over a specific time period, is represented as a linear optimization problem applied to a collection of open stations. With operating capital, the owner of the system can open new stations. This property introduces a set-dependent aspect to the duration required for opening a new station, and the optimal investment problem can be viewed as a variant of the Traveling Salesman Problem (TSP) with set-dependent cost. We propose an A* search algorithm to address this particular variant of the TSP. Computational experiments highlight the benefits of the proposed algorithm in comparison to the widely recognized Dijkstra algorithm and propose future research to explore new possibilities and applications for both exact and approximate A* algorithms.

\keywords{Autonomous
Mobility on-demand  \and vehicle-sharing \and traveling salesman problem \and A* algorithm}
\end{abstract}

\section{Introduction}
Mobility on demand (MoD) is a rapidly growing market\footnote{https://www.alliedmarketresearch.com/mobility-on-demand-market}. With the advanced technology of autonomous vehicles, Autonomous Mobility on demand (AMoD) is becoming increasingly popular because it alleviates some operational difficulties of MoD. The global market for autonomous mobility is projected to grow from 5 billion USD (in $2019$) to $556$ billion USD (in $2026$)\footnote{https://www.alliedmarketresearch.com/autonomous-vehicle-market}, promising safety ($94\%$ of accidents caused by human factors), increased performance, improved efficiency, and more affordable services.

Although auto manufacturers and major technology firms have the resources to quickly establish an AMoD system, smaller operators of shared mobility and public authorities may encounter challenges in securing enough initial capital to launch the service with a sufficient fleet\footnote{https://www.weforum.org/agenda/2021/11/trends-driving-the-autonomous-vehicles-industry/}. Consequently, small companies start to operate in a smaller region, as studied in the literature on optimal service region design, e.g., \cite{he2017service,hao2021prohibiting}. As operators accumulate profits, they can gradually acquire more vehicles and expand their active sites. This research considers such a \textit{refinancing} model of the AMoD system, where the operator aims to achieve the desired service area and the size of the fleet as quickly as possible. 

The existing literature covers a spectrum of topics related to AMoD systems, including aspects such as vehicle-sharing system operations, strategic decision-making, and regulatory and subsidy considerations. Relevant sources can be found in works such as \cite{braverman2019empty, freund2018minimizing, george2011fleet, hao2021prohibiting}. To the best of our knowledge, the question of what is the optimal investment sequence to build an AMoD has not been addressed yet. In this research, we consider an AMoD with a target service area, as well as a current set of open stations. The operator decides on the sequence in which they open the stations. The more profit they make, the faster they can open new stations.

In the following sections, we address the above questions and then analyze the performance of our proposed algorithm. To do so, we review publications close to our research in Section 2. Next, we present the problem statement and related formulations in Section 3. Section 4 presents our solution approach based on the A* search algorithm. Numerical experiments and some promising results are presented and analyzed in Section 5. Finally, Section 6 concludes and points out further research directions based on the current research.

\section{Literature}\label{sec:literature}


This section provides a brief literature review on AMoD systems. Research into the operation and planning of AMoD systems encompasses a range of questions. However, its main emphasis lies in optimizing an existing vehicle-sharing network. Regarding fleet optimization, we can refer to \cite{george2011fleet, nair2011fleet, freund2018minimizing} and \cite{lu2018optimizing}. George and Xia \cite{george2011fleet} study a fleet optimization problem in a closed queue network. This work suggests basic principles for the design of such a system. Nair and Miller-Hooks \cite{nair2014equilibrium} use the equilibrium network model to find the optimal configuration of a vehicle-sharing network. The solutions to the model explain the correctness of the equilibrium condition, the trade-offs between operator and user objectives, and the insights regarding the installation of services. Freund et al. \cite{freund2018minimizing} address how to (re-)allocate dock capacity in vehicle-sharing systems by presenting mathematical formulations and a fast polynomial-time allocation algorithm to compute an optimal solution. Lu et al. \cite{lu2018optimizing} consider the problem of allocating vehicles to service zones with uncertain one-way and round-trip rental demand. 

Regarding policies, Martin et al. \cite{martin2020value} conclude that the use of driverless vehicles and human-driven vehicles can improve profits and operators can gain new unprofitable markets for them. The authors propose a model and an algorithm to find maximum profit while considering driverless and human-driven vehicles. Hao and Martin \cite{hao2021prohibiting} present a model that studies the impact of regulations on the decisions of vehicle-sharing operators and measures the efficiency and effectiveness of these regulations. The results show that the interdependencies between regulations and societal welfare indicators are non-trivial and possibly counterintuitive.

To conclude, we observe that all the research so far has tried to address different questions with the goal of optimizing an already established vehicle-sharing network. However, the question of how to establish new stations and acquire new vehicles has not been addressed yet. In the following, we introduce an optimization problem aimed at identifying the optimal sequence for station establishment and the fleet size required to reach the end state where all stations are operational in the shortest possible time.

\section{Problem Statement and Formulation}
We study an optimal investment strategy for an AMoD (Autonomous Mobility-on-Demand) operator to increase their fleet size and operating area. The AMoD operator's business area comprises stations, $(\mathcal{R}: \{1, ..., R\})$. \enquote{Station} can also refer to a virtual location, e.g., the center of a region in a free-floating system. The operating station $i$ incurs an initial cost $c^b_i$ related to construction, permits, or marketing. Some stations are already open, and profits will be collected from already open stations to increase the budget for new stations. The operator incrementally grows the fleet to reach the optimal size promptly while ensuring acceptable service levels within a gradually expanding operating area.
At a given open station $i$, customers begin their journeys to a different station $j$. When a station is not operational, customers intending to start or complete their journeys there can opt for a neighboring station. 
Customer arrivals are modeled by a Poisson distribution with an arrival rate denoted as $\lambda_{ij}$, and 0 when at least on of the stations is closed. 
The travel times between the stations are exponentially distributed, with an average of $1/\mu_{ij}$, where $\mu_{ij}$ denotes the return rate. These arrival and return rates remain constant and are determined solely by whether stations $i$ and $j$ are open.

The operator determines the fleet size $n$ at any given time, allowing it to grow during expansion. Each new vehicle acquisition comes with a procurement cost of $c^p$. The fleet size must be large enough to serve at least a fraction $\alpha$ of all customers, meeting the minimum service level requirement for the AMoD system. Throughout the development of the AMoD service, it is crucial to keep the service level constant to offset the potential learning effects that could deter customers from using the service \cite{decroix2021service}. To maintain the service level, the operator can rebalance vehicles between stations, incurring a cost of $c^r_{ij}$. The operator receives a contribution margin of $\delta_{ij}$ for each served customer traveling from station to station, representing the payoff minus direct operating costs such as fuel and periodic repairs.

Consequently, this problem involves two decision-making components: establishing the optimal investment plan, which includes timing, locations, and quantity for opening new stations and vehicle acquisition, and overseeing fleet operations, which includes vehicle rebalancing. The model for determining the optimal fleet size and an algorithm for determining investment sequence are introduced in the subsequent sections.

\subsection{Semi-Markov Decision Process for Determining the Optimal Fleet Size}
We see the optimal investment scheduling problem of AMoD operators as a semi-Markov decision process (SMDP) due to the nature of the investment problem. In an SMDP, the system's state evolves according to a semi-Markov process, and the decision-maker selects actions based on the current state. \\
\textbf{States}
Each state $s\in\mathcal{S}$ describes the current fleet of size $n$ and the currently open stations, given by $x_i = 1$ if station $i\in\mathcal{R}$ is open, $0$ otherwise. 
\begin{align*}
s = \langle n, x_1, \dots, x_{R}\rangle
\end{align*}





Each state $s$ is associated with an operational profit $p(s)$ per period, which is calculated by subtracting the rebalancing costs from the contribution margins and an acquisition cost $c(s)$ related to the procurement cost of all vehicles and the cost incurred due to the opening of the station. Apparently, we only need to consider states with positive operational profit in our investment scheme. Regarding this point, the set of states with positive operational profit and the starting state is denoted as $\mathcal{S}$. Also, if a state $s'$ contains all open stations in a state $s$, we can easily see and prove that $p(s')\geq p(s)$. For referencing the fleet size and open stations of a specific state $s$, the notation $n\left(s\right)$ and $x_i\left(s\right)$ are utilized, respectively. Then, the value of $c(s)$ is determined as follows:
\begin{align*}
    c(s) = n(s)\cdot c^p + \sum_{i\in \mathcal{R}}x_i(s) c^b_i
\end{align*}
\textbf{Actions}
Actions refer to the operator's procurement decision, resulting in a state transition to the target state $t\in\mathcal{S}$. Every state $s\in\mathcal{S}$ allows transitions to all other states such that no stations are being closed, that is, $s\rightarrow t$ exists if $x_i\left(s\right) \leq x_i\left(t\right) \forall i$. 

The time $\tau(s,t)$ necessary for a state transition from state $s$ to a state $t$ depends on the operational profit $p(s)$ and the necessary investment volume $C(s,t)$ where 
\begin{align*}
C(s,t) = c(t) - c(s) = \left(n\left(t\right) - n\left(s\right)\right)\cdot c^p + \sum_{i \in \mathcal{R}}\left(x_i(t) - x_i\left(s\right)\right) \cdot c^b_i.
\end{align*}
Given that we do not consider partial states (e.g., a state without optimal fleet size), this means that $p(s)$ is considered the maximum profit corresponding to state $s$, and the optimal decision is to transition to the next state as soon as possible. Thus, \(\tau(s,t) = \frac{C(s,t)}{p(s)}\).

We notice that if $|t|\geq |s|+2$, it is more advantageous to transition to an immediate state $s'$ where $|s|< |s'|<|t|$ because $\frac{C(s,t)}{p(s)} \geq \frac{C(s,s')}{p(s)} + \frac{C(s',t)}{p(s')}$ due to the fact that $p(s) \leq p(s')$ and $C(s,t) = C(s,s') + C(s',t)$. Therefore, we only need to consider actions between two consecutive states in any optimal investment scheme.

\subsection{A model for calculating optimal profit and minimum acquisition cost}
To compute the operational profit $p(s)$ per state $s\in\mathcal{S}$, we formulate the rebalancing problem as an open-queueing network (in line with, e.g., \cite{braverman2019empty,hao2021prohibiting, he2017service,martin2020value}), and optimize over it to maximize operational profits. Given a set of available stations, the model determines the necessary size of the fleet to reach the level of service and rebalance. Since we want to maximize profit and minimize the corresponding acquisition cost, our objective function is hierarchical since we optimize the second objective after minimizing the first objective.

To start, we denote $f_{ij}, e_{ij}$ ($i\neq j$) as the number of occupied and empty vehicles traveling from $i$ to $j$ and $e_{ii}$ as the number of idle vehicles currently parked at station $i$.
To determine the maximum operational profit per period for state $s$, we solve \eqref{eq:objectiveRebalancing} - \eqref{eq:domain1} for all opening stations in $R_s = \{i\in \mathcal{R} | x_i\left(s\right) = 1\}$. The mathematical formulation is expressed as follows:
\begin{multline}
    P(obj_1,obj_2) = \Bigl(\max \,\,\, \alpha \Bigl( \sum_{i\in R_s} \sum_{j\in R_s}{\lambda_{ij}\delta_{ij}} \\ -\sum_{i\in R_s}\sum_{j\in R_s}{c^r_{ij}\mu_{ij}e_{ij}} \Bigl),\min \,\,\,  \Bigl( n\cdot c^p+\sum_{i\in R_s}c^b_i \Bigl) \Bigl)
    \label{eq:objectiveRebalancing}
\end{multline}
subject to
\allowdisplaybreaks
\begin{align}
\lambda_{ij} &=\mu_{ij}f_{ij}, &\forall i,j\in R_s\label{eq:fullCarLittlesLaw} \\
\sum_{j\in R_s \setminus\{i\}}\mu_{ji}e_{ji} &\leq \sum_{j\in R_s \setminus\{i\}}\lambda_{ij},  &\forall i\in R_s\label{eq:emptyCarLittlesLaw} \\
\sum_{j\in R_s}\lambda_{ij}+\sum_{j\in R_s}\mu_{ij}e_{ij}&=   \sum_{j\in R_s}\mu_{ji}e_{ji} + \sum_{j\in R_s}\lambda_{ji}, &\forall i\in R_s\label{eq:carFlowBalance2}\\ 
\frac{\alpha}{1-\alpha} &\leq e_{ii}, &\forall i\in R_s\label{eq:actualavailability2} \\
\sum_{i,j\in R_s}{\left(e_{ij}+f_{ij}\right)}&=n,\label{eq:fleetDimension} \\
e_{ij}, f_{ij}  &\geq 0, &\forall i,j\in R_s\label{eq:domain1}
\end{align}
 

The objective function \eqref{eq:objectiveRebalancing} maximizes profit by dividing the contribution margin of all served customers by rebalancing costs, multiplied by availability $\alpha$, and minimizing set-up fees. Constraints \eqref{eq:fullCarLittlesLaw} - \eqref{eq:carFlowBalance2} linearize flow constraints in queueing networks, almost directly follow from \cite{braverman2019empty} and requiring the system to achieve a service level of at least $\alpha$, eliminating any upper bound on demand, unlike \cite{braverman2019empty}. Constraints \eqref{eq:actualavailability2} set the required safety stock, following the fixed population mean approximation in open queueing networks due to \cite{whitt1984open}. Constraints \eqref{eq:fleetDimension} bound fleet size, and constraints \eqref{eq:domain1} defined the domain.

\section{Solution Approach}

It is important to note that in our problem, the optimal time for opening a new station depends on profits from existing stations, resulting in a set-dependent cost. The exponential growth of these sets makes mathematical representations potentially too complex, making contemporary solvers unsuitable for modeling and solving this formulation.

We can consider the investment problem as a variant of the well-known Traveling Salesman Problem (TSP) with set-dependent travel costs. Taking into account a permutation $(u_1,u_2,..,u_n)$ that presents an order that the stations are opened. Each subpath $(u_1,u_2,..,u_i)$ is assigned a state $s_i$ where $x_k(s_i)=1$ if $u_j=k$ for some $j=1..i$. The cost between two consecutive states, $s_i$ and $s_{i+1}$, is calculated using the formulations in Section 3.1, which depend on the set of open stations in $s_i$. In other words, it is a set-dependent cost function. While there is much research for TSP in general and several studies on level-dependent travel cost TSP \cite{bigras2008time,alkaya2015combining} in particular (the cost associated with each city depends on the index of that city in the solution), our cost function makes the problem cannot be modeled with formulations similar to the ones for TSPs. 

\subsection{Heuristic strategy for A* algorithm}
We model our investment problem as a shortest path problem. Consider a graph $G=(V,A)$ where each node $n_s\in V$ corresponds to the state $s$. Each arc $(n_s,n_{s'})\in A$ corresponds to a feasible action between two consecutive states $s$ and $s'$ with cost $C(s,s')$. Finding the shortest investment time is equivalent to finding the shortest path from node $n_{s_0}$ to node $n_{s_f}$ where $s_0$ and $s_f$ are the initial state and the final state, respectively. Since we can define a 1-1 mapping between $s$ and $n_s$, we subsequently use $s$ instead of $n_s$ to simplify the notation.

 
To solve this shortest-path problem, we rely on the A* algorithm. Given a state $s$, unlike the classic Dijkstra algorithm, which only evaluates the cost of the shortest path $g(s)$ from the source $s_0$ to $s$, A* also evaluates the cost $h(s)$ from $s$ to the final state $s_f$, and the cost for each node $s$ is then $f(s)=g(s) + h(s)$ instead of $g(s)$. The A * algorithm can always find the shortest path from $s_0$ to $s_f$ if $h(s)$ does not exceed the cost of the shortest path from $s$ to $s_f$ for any $s$. Otherwise, A* becomes a heuristic algorithm. \\
\textbf{Simple heuristic for A*}
We start with some of the simplest heuristics for A*. Given that the current, next, and final states are $s, s'$ and $s_f$, the cost of the shortest path from $n_0$ to $s'$, $g(s')$, is $g(s') = g(s) + \frac{c(s') - c(s)}{p(s)}$. Several simple ways to calculate $h(s')$ are as follows (where $eh$ and $ah$ denote exact and approximate heuristics, respectively):
\begin{align}
eh_1\left(s'\right) &=  \frac{c(s_f) - c(s')}{P_{R-1}} \label{eq:underestimate} \\ 
ah_1\left(s'\right) &=  \frac{c(s_f) - c(s')}{p(s')} \label{eq:overestimate}
\end{align}
Heuristic functions \eqref{eq:underestimate}, \eqref{eq:overestimate} underestimate and overestimate the shortest time of the optimal path from $s_0$ to $s_f$ that passes through $s'$. Here, $P_{R-1}$ denotes the maximum profit for any state that has $R-1$ open stations. Using a linear combination, we obtain other heuristics where $\gamma \in [0,1]$ is a parameter that can be a fixed constant or dynamically adjusted during the execution of the algorithm. We aim to test whether we can obtain simple heuristics that may not be optimal but can quickly find reasonable solutions.
 \begin{align}
ah_2\left(s'\right) &=  \gamma eh_1(s') + (1-\gamma)ah_1(s') \label{eq:combination1} 
\end{align}
\textbf{Stronger lower bound  heuristics for A*} Assume that $s=s_1$ is the current state. Let $s_1,s_2,..,s_k$ be a sequence of states where $s_{i+1}$ is obtained from $s_i$ by adding a new station and $s_k = s_f$ be the final state where all stations are open. The total transition time from state $s_1$ to state $s_k$, $\tau(s_1,..,s_k)$, is:
\begin{equation}
\tau(s_1,..,s_k) = \frac{c(s_2) - c(s_1)}{p(s_1)} + \frac{c(s_3) - c(s_2)}{p(s_2)} + \ldots + \frac{c(s_k) - c(s_{k-1})}{p(s_{k-1})}\label{eq:oriEq}
\end{equation}
We denote $\underline{\Delta_{c}}(s_i)$ as a lower bound of the difference of the acquisition cost $c(s_{i+1}) - c(s_{i})$ between two consecutive states $s_i$ and $s_{i+1}$. Let $P_m$ be a state with the maximum profit among all states with $m$ opening stations. We can find the value of $P_m$ by solving the model (\ref{eq:objectiveRebalancing1}) - (\ref{eq:domain2})
(see Appendix), which aims to maximize the profit and minimize the corresponding acquisition cost given a fixed number of stations that can be opened. Then, we obtain $p(s_i) \leq P_{|s_i|}, \forall i=1,\ldots,k$. The values of $P_{|s_i|}$ define an increasing sequence since we open more stations. Therefore, we have
$p(s_i) \leq P_{|s_i|} \leq P_{|s_{k-1}|} = P_{R-1}, \forall i = 1,\ldots,k-1$.
Given that $c(s_{i+1}) - c(s_{i}) \geq \underline{\Delta_{c}}(s_i)$ or $c(s_{i+1}) - c(s_{i}) - \underline{\Delta_{c}}(s_i) \geq 0$, therefore, $\forall i = 1,\ldots,k-1$ we have the following.
\begin{align}
\frac{c(s_{i+1}) - c(s_i)}{p(s_i)} 
 &=
 \frac{\underline{\Delta_{c}}(s_i) + (c(s_{i+1}) - c(s_i) - \underline{\Delta_{c}}(s_i))}{p(s_i)} \\
 &= \frac{\underline{\Delta_{c}}(s_i)}{p(s_i)} +  \frac{c(s_{i+1}) - c(s_i) - \underline{\Delta_{c}}(s_i)}{p(s_i)} \\
&\geq 
 \frac{\underline{\Delta_c}(s_i)}{P_{|s_i|}} 
+ \frac{c(s_{i+1}) - c(s_i) - \underline{\Delta_c}(s_i)}{P_{R-1}} 
\end{align}
and consequently:
\begin{align}
\tau(s_1,\ldots,s_k)\geq 
 \left(\sum_{i=1}^{k-1}\frac{\underline{\Delta_c}(s_i)}{P_{|s_i|}}\right) + \frac{c(s_{k}) - c(s_{1}) - \sum_{i=1}^{k-1}\underline{\Delta_c}(s_i)}{P_{R-1}}\label{eq:firstLB}
\end{align}
Inequality \eqref{eq:firstLB} gives us a more robust lower bound than the simple one presented in \eqref{eq:underestimate}. \\
\textbf{Evaluating the lower bound $\underline{\Delta_c}(s_1)$}
From (\ref{eq:fleetDimension}), we see that with each state $S$, the optimal number of vehicles $n$ is equal to $\sum_{i,j\in S}(e_{ij}+f_{ij})$. Following $obj_2$, when we open a new station, the acquisition cost includes the cost of station setup and the new vehicle acquisition cost. We assume that the difference in acquisition cost between two consecutive states depends on the values $f_{ij}, e_{ij}$ of the new station $i$. With this assumption, the minimum acquisition cost of opening station $i$ from a given state $S$ ($i\notin S$) to obtain maximum profit is determined by $\Delta_c(S, i)$. In other words, $\Delta_c(S, i)$ presents the lower difference in acquisition cost between two consecutive states in which the next state is reached by opening station $i$ from the state $S$.
\begin{align}
\Delta_c(S, i) = \sum_{j\in S}(e_{ij} + e_{ji} + f_{ij} + f_{ji})c^p + c_b^i \quad \forall i\notin S\label{eq:minAcquisitionI}
\end{align}
Then, $c(s_{i+1}) - c(s_{i}) \geq \min_{o\notin s_{i}}\Delta_c (s_i,o)\quad \forall i = 1,2,\ldots,k-1$. 

Next, we show how to obtain the lower bound $\Delta_c (s_i,o)$ using equations \eqref{eq:minAcquisitionI}. Assuming $T\subset \mathcal{R}$ be the state with any $t$ stations not in $S$, $t < R-|S|, S\cap T = \emptyset$. Let $o\notin S \cup T$, and we evaluate $\Delta C(S, t, i)$ - the minimum acquisition cost difference when building a new station $i$ starts from state $S\cup T$ with any state $T$ such that $|T| = t$.\\
\textbf{Underestimate acquisition cost}
We rewrite $\Delta C(S,t,i)$ using equation \eqref{eq:minAcquisitionI} as follows:
\begin{equation}
    \Delta C(S,t,i) = \min_{T\subset \mathcal{R}, |T| = t}\sum_{j\in S \cup T}(e_{ij} + e_{ji} + f_{ij} + f_{ji})c^p + c^b_i
\end{equation}
and since $e_{ij}, e_{ji}$ and $c^p$ are non-negative, we have
\begin{equation}
    \Delta C(S,t,i) \geq \min_{T\subset \mathcal{R}, |T| = t}\sum_{j\in S \cup T}(f_{ij} + f_{ji})c^p + c^b_i
\end{equation}
Since $c^b_i + \sum_{j\in S}(f_{ij}+f_{ji})c^p$ is a constant, we will develop a lower bound for the sum $\sum_{j\in T}(f_{ij} + f_{ji})c^p$. Apparently, $\sum_{j\in T}(f_{ij}+f_{ji})c^p$ cannot be smaller than  the sum of $|T|$ smallest values of $(f_{ij} + f_{ji})c^p$ where $j\notin S \cup \{i\}$. Therefore, we developed the Algorithm \ref{alg:cap1} to evaluate a lower bound of $\Delta C(S,t,i)$.
\begin{algorithm}[!ht]
\caption{Lower bound evaluation of acquisition cost}\label{alg:cap1}
\begin{algorithmic}[1]
\Require $S, i, t$
\Ensure A lower bound of $\Delta C(S,t,i)$
\State Let $\alpha \leftarrow c^b_i + \sum_{j\in S}(f_{ij}+f_{ji})c^p$
\State Sort ${(f_{ij} + f_{ji})}_{j\in \mathcal{R}\setminus (S \cup \{i\})}$  increasingly. 
\State Let $(f_{ij_1} + f_{j_1i})  \leq (f_{ij_2} + f_{j_2i}) \leq ... \leq (f_{ij_k} + f_{j_ki}) \leq ...$ be the array after sorting.
\State Let $\beta \leftarrow \sum_{k=1}^t(f_{ij_k} + f_{j_ki})c^p$
\State \textbf{Return} $\alpha + \beta$ 
\end{algorithmic}
\end{algorithm}
\\Using Algorithm \ref{alg:cap1}, we have that
\begin{equation}
\Delta_c(s_i,o) \geq \Delta C(s_1, i-1, o) \geq c^b_o + \sum_{j\in s_1}(f_{oj}+f_{jo})c^p + \sum_{k=1}^{i-1}(f_{oj_k} + f_{j_ko})c^p
\end{equation}
and consequently,
\begin{align}
c(s_{i+1}) - c(s_{i}) &\geq \min_{o\notin s_i} \Delta_c(s_i,o) \\
                      &\geq \min_{o\notin s_i} \left(c^b_o + \sum_{j\in s_1}(f_{oj}+f_{jo})c^p + \sum_{k=1}^{i-1}(f_{oj_k} + f_{j_ko})c^p\right)\label{eq:stronger_lb}
\end{align}
Use $\underline{\Delta_c}(s_i) = \min_{o\notin s_i} \left(c^b_o + \sum_{j\in s_1}(f_{oj}+f_{jo})c^p + \sum_{k=1}^{i-1}(f_{oj_k} + f_{j_ko})c^p\right)$ in inequality (\ref{eq:firstLB}), we obtain a lower bound heuristic for A*, called $eh_2$, which is stronger than the simple one $eh_1$. However, we need to solve it online.

We obtain a weaker lower bound version of $eh_2$ by fixing $s_1$, e.g., to the initial state $s_0$. Still, this strategy may reduce the total running time since the value of $\underline{\Delta_c}(s_i)$ needs to be calculated only once, while with $eh_2$, we will calculate $\underline{\Delta_c}(s_i)$ for each extracted state $s=s_1$ from the queue. Let $eh_3$ be this lower bound heuristic.

\section{Numerical Experiments}

In this section, we present the numerical design and then report the experiment results of exact and heuristic algorithms to find an optimal schedule investment. The algorithm and formulations were written in C++, and the MILP models were solved by CPLEX 22.1.1. The experiments were run on an AMD Ryzen 3 3100 machine with a 4-core processor, 3.59 GHz, and 16GB of RAM on a 64-bit Windows system.
\subsection{Numerical design}
We conducted experiments on randomly generated datasets, following a similar approach as Martin et al. \cite{hao2021prohibiting}. To model the real-world transportation network structure, our datasets vary in size ($R\in \{7,9,16,19,25\}$ and geographic distributions of station locations, including circular (C), hexagonal (H), and quadratic (Q) layouts. The methodology for generating data and configuring model parameters is elucidated in the Appendix.


The investment starts with a set of initially open stations. We assume that initially, there is a budget of $B=10000$, optimally utilized to construct the initial stations to maximize the initial profit. With smaller instances (less than $10$ stations), we use a dynamic budget of $500\times R$ to avoid opening too many stations in the initial state. We simulate this process through formulations similar to (\ref{eq:objectiveRebalancing1}) - (\ref{eq:domain2}) (see Appendix). With this budget, the initial state has $5-7$ open stations for larger instances and $2-3$ for instances with fewer than 10 stations. Then, the A* algorithm will find an optimal investment plan starting from the initial state with a certain number of already opened stations obtained from the formulations.

\subsection{Results}
In the following, we assess the following two points:
\begin{enumerate}
    \item We compare the performance of the exact A* heuristics and Dijkstra algorithm based on the execution time, the number of states explored, and the number of states remaining in the priority queue.
    \item We compare the performance of approximate A* heuristics in terms of optimal gap and execution time.
\end{enumerate}
Table \ref{tbl:exactA*} analyzes the performance of the exact A * algorithms and the Dijkstra algorithm by reporting their running time in seconds (column Time (s)), the number of nodes extracted by the A* algorithm (column Exp.), and the number of nodes still in the queue (column Rem.) with the optimal value (column Opt.) obtained from all exact algorithms.

\begin{table}[!ht]
\caption{Results of exact A* heuristic and Dijkstra algorithms} \label{tbl:exactA*}
\begin{center}
\addtolength{\leftskip} {-2cm}
\addtolength{\rightskip}{-2cm}
\begin{tabular*}{1.2\textwidth}{@{\extracolsep{\fill}}l>{\raggedleft}p{0.08\linewidth}>{\raggedleft}p{0.09\linewidth}>{\raggedleft}p{0.09\linewidth}>{\raggedleft}p{0.09\linewidth}>{\raggedleft}p{0.09\linewidth}>{\raggedleft}p{0.09\linewidth}>{\raggedleft}p{0.09\linewidth}>{\raggedleft}p{0.09\linewidth}>{\raggedleft}p{0.09\linewidth}>{\raggedleft}p{0.09\linewidth}r}
\hline
\multirow{2}{*}{Instance} & \centering\multirow{2}{*}{Opt.} & \multicolumn{1}{c}{Dijkstra} & \multicolumn{3}{c}{A* + $eh_1$} & \multicolumn{3}{c}{A* + $eh_2$} & \multicolumn{3}{c}{A* + $eh_3$} \\ \cmidrule(lr){3-3} \cmidrule(lr){4-6} \cmidrule(lr){7-9}  \cmidrule(lr){10-12}
 ~ & ~ & \multicolumn{1}{c}{Time (s)} & \multicolumn{1}{c}{Exp.} & \multicolumn{1}{c}{Rem.} & \multicolumn{1}{c}{Time (s)} & \multicolumn{1}{c}{Exp.} & \multicolumn{1}{c}{Rem.} & \multicolumn{1}{c}{Time (s)} & \multicolumn{1}{c}{Exp.} & \multicolumn{1}{c}{Rem.} & \multicolumn{1}{c}{Time (s)} \\ \hline
 C-7-BAL & 1563.19 & <1 & 17 & 144 & <1 & 12 & 10 & <1 & 12 & 12 & <1 \\ 
 H-7-BAL & 1524.71 & <1 & 17 & 13 & <1 & 11 & 9 & <1 & 12 & 12 & <1 \\ 
 Q-9-BAL & 435.53 & <1 & 33 & 31 & <1 &  14 & 25 & <1 &  20 & 25 & <1 \\ 
 Q-16-BAL & 420.87 & 1 & 392 & 173 & 1 & 227 & 236 & 1 & 243 & 231 & 1 \\
 Q-16-IMB & 723.69 & 1 & 340 & 176 & 1 & 150 & 208 & 1 & 170 & 195 & 1 \\ 
 C-19-BAL & 1054.83 & 14 & 2733 & 2060 & 12 & 1453 & 2234 & 9 & 1603 & 2156 & 9 \\
 C-19-IMB & 1681.48 & 14 & 2292 & 1473 & 11 & 902 & 1421 & 7 & 1103 & 1378 & 7 \\
 H-19-BAL & 1028.32 & 13 & 2303 & 1710 & 11 & 903 & 1568 & 7 & 1149 & 1656 & 8 \\ 
 H-19-IMB & 1833.02 & 14 & 1812 & 1720 & 10 & 592 & 1299 & 5 & 888 & 1385 & 6 \\ 
 Q-25-BAL & 711.31 & 1313 & 140878 & 119407 & 1032 & 35068 & 84174 & 452 & 44551 & 90771 & 508 \\ 
 Q-25-IMB & 1185.37 & 1311 & 124532 & 105743 & 978 & 18440 & 55015 & 328 & 24975 & 61453 & 372\\ \hline
    \end{tabular*}
    \end{center}
\end{table}

The experiments show that datasets with imbalanced arrival rates take longer to open stations due to decreased profit margins. The strongest lower bound heuristic, $eh_2$, has the shortest running time, number of expanded nodes, and number of remaining nodes among all exact methods, detailed in Table \ref{tbl:exactA*}. Using the A* algorithm with $eh_2$ significantly reduces computation time and vertice exploration compared to underestimating the optimal path's shortest time $eh_1$. The exact heuristic $eh_3$ also provides computational stability without online updates.

We observe that the number of visited vertices and execution time increases exponentially with the number of stations. To find a suitable investment schedule, the researchers experimented with various heuristic approximation approaches in the A* search algorithm. Results in Table \ref{tbl:approximateA*} showed that larger values of $\gamma$ resulted in better objective values and longer running time. Although these approaches achieve excellent time efficiency and small gaps, they are highly dependent on data and can become less effective when parameter ranges are modified.

Finally, we report the performance of weighted A* variants in Table \ref{tbl:approximateA*1}, which multiply the values of $eh_2$ and $eh_3$ by $1.05$ or $1.1$. The best solutions ensure a gap between optimal and best solutions of at most $5\%$ or $10\%$. Although slower than the ones mentioned in Table \ref{tbl:approximateA*}, it ensures an optimal gap that the approximate heuristics cannot. The results show that the optimal gap obtained by these approximation algorithms is very small, highlighting the effectiveness of both heuristics.
\begin{table}[H]
\centering
\label{tbl:approximateA*}
\caption{Non-bounded approximation algorithms with simple heuristics.}
\begin{tabular}{l>{\raggedleft}p{0.08\linewidth}>{\raggedleft}p{0.09\linewidth}>{\raggedleft}p{0.09\linewidth}>{\raggedleft}p{0.09\linewidth}>{\raggedleft}p{0.09\linewidth}>{\raggedleft}p{0.09\linewidth}>{\raggedleft}p{0.09\linewidth}>{\raggedleft}p{0.09\linewidth}r}
\hline
 \multirow{2}{*}{Instance} & \centering\multirow{2}{*}{Opt.} & \multicolumn{2}{c}{A* + $ah_1$}  & \multicolumn{2}{c}{A* + $ah_2(\gamma = 0.3)$}   & \multicolumn{2}{c}{A* + $ah_2(\gamma = 0.5)$}   & \multicolumn{2}{c}{A* + $ah_2(\gamma = 0.7)$}   \\ \cmidrule(lr){3-4} \cmidrule(lr){5-6} \cmidrule(lr){7-8} \cmidrule(lr){9-10} 
 ~ & ~ & Gap (\%) & Time (s) & Gap (\%) & Time (s) & Gap (\%) & Time (s) & Gap (\%) & Time (s) \\ \hline
 C-7-BAL & 1563.19 & 9.01 & <1 & 4.34 & <1 & 0.00 & <1 & 0.00 & <1 \\ 
 H-7-BAL & 1524.71 & 8.85 & <1 & 4.37 & <1 & 4.37 & <1 & 0.00 & <1 \\ 
 Q-9-BAL & 435.53 & 4.63 & <1 & 1.29 & <1 & 1.29 & <1 & 0.00 & <1 \\ 
 Q-16-BAL & 420.87 & 7.18 & <1 & 3.86 & <1 & 1.78 & <1 & 0.00 & 1 \\ 
 Q-16-IMB & 723.69 & 6.00 & <1 & 3.82 & <1 & 1.47 & <1 & 0.00 & 1 \\ 
 C-19-BAL & 1054.83 & 12.76 & <1 & 7.78 & <1 & 0.94 & <1 & 0.00 & 3 \\ 
 C-19-IMB & 1681.48 & 17.25 & <1 & 9.62 & <1 & 0.33 & <1 & 0.00 & 3 \\ 
 H-19-BAL & 1028.32 & 6.40 & 1 & 2.11 & <1 & 1.20 & <1 & 0.00 & 2 \\ 
 H-19-IMB & 1833.02 & 11.98 & <1 & 9.19 & <1 & 4.48 & <1 & 0.00 & 1 \\
 Q-25-BAL & 711.31 & 15.10 & 1 & 10.08 & 1 & 6.84 & 1 & 0.53 & 39 \\ 
 Q-25-IMB & 1185.37 & 11.87 & 1 & 7.28 & 1 & 3.39 & 1 & 1.26 & 15 \\ \hline
    \end{tabular}
\end{table}

\begin{table}[!ht]
\caption{Bounded approximation algorithms based on stronger lower-bound heuristic}\label{tbl:approximateA*1}
\begin{tabular}{l>{\raggedleft}p{0.08\linewidth}>{\raggedleft}p{0.09\linewidth}>{\raggedleft}p{0.09\linewidth}>{\raggedleft}p{0.09\linewidth}>{\raggedleft}p{0.09\linewidth}>{\raggedleft}p{0.09\linewidth}>{\raggedleft}p{0.09\linewidth}>{\raggedleft}p{0.09\linewidth}r}
\hline
\multicolumn{1}{c}{\multirow{2}{*}{Instance}} & \multicolumn{1}{c}{\multirow{2}{*}{Opt.}} & \multicolumn{2}{c}{A* + $1.1*eh_2$}  & \multicolumn{2}{c}{A* + $1.1*eh_3$}  & \multicolumn{2}{c}{A* + $1.05*eh_2$} & \multicolumn{2}{c}{A* + $1.05*eh_3$} \\\cmidrule(lr){3-4} \cmidrule(lr){5-6} \cmidrule(lr){7-8} \cmidrule(lr){9-10} 
\multicolumn{1}{c}{}  & \multicolumn{1}{c}{}      & \multicolumn{1}{c}{Gap (\%)} & \multicolumn{1}{c}{Time (s)} & \multicolumn{1}{c}{Gap (\%)} & \multicolumn{1}{c}{Time (s)} & \multicolumn{1}{c}{Gap (\%)} & \multicolumn{1}{c}{Time (s)} & \multicolumn{1}{c}{Gap (\%)} & \multicolumn{1}{c}{Time (s)} \\\hline
C-7-BAL  & 1563.19 & 0.00 & \textless{}1 & 0.00 & \textless{}1 & 0.00 & \textless{}1 & 0.00 & \textless{}1 \\
H-7-BAL  & 1524.71  & 0.00 & \textless{}1 & 0.00 & \textless{}1 & 0.00 & \textless{}1 & 0.00 & \textless{}1 \\
Q-9-BAL  & 435.53  & 0.67 & \textless{}1 & 0.67 & \textless{}1 & 0.00 & \textless{}1 & 0.00 & \textless{}1 \\
Q-16-BAL & 420.87 & 0.17 & 1 & 0.13 & 1 & 0.13 & 1 & 0.13 & 1 \\
Q-16-IMB & 723.69 & 0.73 & 1 & 0.00 & 1 & 0.00 & 1 & 0.00 & 1 \\
C-19-BAL & 1054.83 & 0.40 & 6 & 0.09 & 6 & 0.09 & 7 & 0.09 & 8 \\
C-19-IMB & 1681.48 & 0.10 & 4 & 0.04 & 5 & 0.04 & 5 & 0.04 & 6 \\
H-19-BAL & 1028.32 & 0.24 & 3 & 0.07 & 4 & 0.07 & 5 & 0.01 & 6 \\
H-19-IMB & 1833.02 & 0.27 & 4 & 0.07 & 4 & 0.14 & 4 & 0.00 & 5 \\
Q-25-BAL & 711.31 & 0.26 & 194  & 0.08 & 241  & 0.04 & 306  & 0.03 & 370  \\
Q-25-IMB & 1185.37 & 0.43 & 117  & 0.09 & 135  & 0.10 & 181  & 0.00 & 241 \\\hline
\end{tabular}
\end{table}

To conclude the section, we observe that for those benchmark instances, exact methods can provide optimal solutions in a reasonable amount of time for those benchmark instances. The proposed lower bound heuristic $eh_2$  beats simple heuristic $eh_1$ and the Dijkstra algorithm. The simple approximate A* heuristic can give quite good results with a small computation time, while the weighted A* heuristic based on the best lower bound heuristic can reduce the computation time and maintain a small optimal gap. 

\section{Conclusion}
We have studied an investment problem that arises in the context of autonomous mobility on demand systems. Given some already open stations, the question is to determine the optimal sequence of opening the remaining stations to minimize the total opening time. We modeled this investment problem as a Semi-Markov Decision Process and viewed this problem as a variant of the TSP problem, where the cost between two vertices $s$ and $t$ depends on the set of already visited vertices belonging to the path from the source vertex to vertex $s$. This special cost function makes the problem impossible to model and solve with current mixed-integer solver technology. We then developed and solved this new variant using the A* algorithm. The experiment results show that the A* algorithm can reduce by half the running time of the Dijkstra algorithm and a simple, exact A* algorithm. Regarding the approximate A* search, the result shows that we can obtain reasonable solutions with a small computation effort.

It is still a challenging task to solve larger problems. Therefore, we are developing and testing more robust lower-bound heuristics for exact A* search. Also, we are testing new approximate heuristics for A* search that take ideas from the lower bound heuristics. The initial results show that we can solve larger instances in a shorter time using both methods. Also, the approximate A* heuristic gives similar results to those returned by the exact A* heuristic in many problem instances.
\section*{Acknowledgement}
The work has been carried out partly at the Vietnam Institute for Advanced Study in Mathematics (VIASM). The corresponding author (Duc Minh Vu) would like to thank VIASM for its hospitality and financial support for his visit in 2023.
\bibliography{ref}
\newpage
\section*{Appendix}
\subsection*{The methodology for generating data and configuring model parameters.}
We randomly sample hourly arrival rates from a uniform distribution $\mathcal{U}[80, 120]$ (BAL). We generate additional instances for larger datasets to reflect imbalances (IMB) in arrival rates. Arrival rates are higher near the city center ($\mathcal{U}[110, 140]$) and lower in suburban areas ($\mathcal{U}[60, 90]$). There are $11$ instances, including $7$ balance instances and $4$ imbalance instances. In all instances, customers travel to other stations with equal probability. Vehicles require a time of $t_{ij} = 3/60 + d_{ij}/25$ hours for the trip from station $i$ to station $j$, where $d_{ij}$ represents the Euclidean distance between the two stations. The minimum level of service for customer retention is $\alpha = 0.5$. Rebalancing a vehicle costs $\$ 0.3 $ per km. Transporting a customer between two locations yields a contribution of $\$0.3$ per kilometer, representing revenues of approximately $\$1$ per kilometer minus direct costs. Procurement costs ($c^p$) are $\$ 1$. The operating cost for each opening station is randomly generated from $\mathcal{U}[1000,3000]$. 
\subsection*{Evaluate profit's upper bound for any state opening $m$ stations}
Assuming that we can open $m$ stations, the following mixed integer formulation (\ref{eq:objectiveRebalancing1}) - (\ref{eq:domain2}). will help us determine the set of stations to open and the number of vehicles to acquire to maximize the profit while minimizing the corresponding acquisition cost.
\begin{multline}
    P(obj_3,obj_4) 
    = \Bigl(\max \,\,\, \alpha \Bigl( \sum_{i\in\mathcal{R}} \sum_{j\in\mathcal{R}}\lambda_{ij}\delta_{ij} \\ -\sum_{i\in\mathcal{R}}\sum_{j\in\mathcal{R}}{c^r_{ij}\mu_{ij}e_{ij}} \Bigl), \min \,\,\, n\cdot c^p+\sum_{i\in\mathcal{R}}c^b_iy_i\Bigl)\label{eq:objectiveRebalancing1}
\end{multline}
\allowdisplaybreaks
\begin{align}
\intertext{subject to}
\mu_{ij}f_{ij} &= \lambda_{ij}x_{ij}, &\forall i,j\in\mathcal{R}\label{eq:flowCons1} \\
\sum_{j\in \mathcal{R} \setminus\{i\}}\mu_{ji}e_{ji} &\leq \sum_{j\in \mathcal{R} \setminus\{i\}}\lambda_{ij}x_{ij},  &\forall i\in \mathcal{R}\label{eq:emptyCarLittlesLawI} \\
\sum_{j\in \mathcal{R}}\lambda_{ij}x_{ij}+\sum_{j\in \mathcal{R}}\mu_{ij}e_{ij}&=   \sum_{j\in \mathcal{R}}\mu_{ji}e_{ji} + \sum_{j\in \mathcal{R}}\lambda_{ji}x_{ji}, &\forall i\in \mathcal{R}\label{eq:carFlowBalance2I}\\
\frac{\alpha}{1-\alpha}\cdot y_i &\leq e_{ii}, &\forall i\in \mathcal{R}\label{eq:actualavailability2I} \\
\sum_{i,j\in \mathcal{R}}{\left(e_{ij}+f_{ij}\right)}&=n,\label{eq:fleetDimensionI} \\
e_{ij} + f_{ij} & \leq Mx_{ij}, &\forall i,j\in\mathcal{R}\label{eq:bigMcons1}\\
x_{ij} &\leq y_i, &\forall i,j\in\mathcal{R}\label{eq:bigMcons2}\\
x_{ij} &\leq y_j, &\forall i,j\in\mathcal{R}\label{eq:bigMcons3}\\
y_{i} + y_{j} - x_{ij} &\leq 1 &\forall i,j\in\mathcal{R}\label{eq:bigMcons4}\\
\sum_{i\in \mathcal{R}}y_i &= m \label{eq:stlimits}\\
e_{ij}, f_{ij}  &\geq 0, &\forall i,j\in \mathcal{R}\label{eq:domain1I}\\
x_{ij}, y_i & \in \{0,1\}, &\forall i,j\in\mathcal{R} \label{eq:domain2}
\end{align}
\subsection*{Strategy for generating initial state given budget B}
Given the budget $B$, the budget is optimally allocated to construct the initial state. The budget limits the number of stations and initial vehicles that can be procured initially, ensuring that the system achieves a predetermined initial profit for development purposes. We can obtain the optimal investment for the initial state using the formulations (\ref{eq:objectiveRebalancing1}) - (\ref{eq:domain2}) and substitute the constraints (\ref{eq:stlimits}) by the budget constraints (\ref{eq:budgetCons}) as follow:

\begin{align}
n\cdot c^p+\sum_{i\in\mathcal{R}}c^b_iy_i & \leq B, &  \label{eq:budgetCons}
\end{align}

\end{document}